\documentclass[letterpaper, 10pt, journal, twoside]{./Style_template/RSS/IEEETrans}  

\usepackage[letterpaper, left=0.7in, right=0.7in, bottom=0.76in, top=0.7in] {geometry}
\def\IEEEtitletopspace{0.2in}
\IEEEoverridecommandlockouts                              
\usepackage{cite}

\usepackage{amsmath}
\usepackage{moreverb,url}
\usepackage{times}
\usepackage{multicol}
\usepackage{graphicx}
\usepackage{amssymb} 
\usepackage{gensymb}
\usepackage{amsmath, xparse}
\usepackage{mathrsfs}
\usepackage{soul}
\usepackage{enumitem,kantlipsum}
\usepackage{multirow}
\usepackage{xcolor}
\usepackage{physics}

\usepackage[colorlinks,bookmarksopen,bookmarksnumbered,citecolor=red,urlcolor=blue]{hyperref}

\newcommand\BibTeX{{\rmfamily B\kern-.05em \textsc{i\kern-.025em b}\kern-.08em
T\kern-.1667em\lower.7ex\hbox{E}\kern-.125emX}}

\def\ie{{\em i.e.,\ }}

\def\vs{{\em v.s.\ }}

\begin{document}

\title{Failure Mechanisms and Risk Estimation for Legged Robot Locomotion on Granular Slopes}
\author{Xingjue Liao and Feifei Qian}

\maketitle

\begin{abstract}

Locomotion on granular slopes such as sand dunes remains a fundamental challenge for legged robots due to reduced shear strength and gravity-induced anisotropic yielding of granular media. Using a hexapedal robot on a tiltable granular bed, we systematically measure locomotion speed together with slope-dependent normal and shear granular resistive forces. While normal penetration resistance remains nearly unchanged with inclination, shear resistance decreases substantially as slope angle increases. Guided by these measurements, we develop a simple robot-terrain interaction model that predicts anchoring timing, step length, and resulting robot speed, as functions of terrain strength and slope angle. The model reveals that slope-induced performance loss is primarily governed by delayed anchoring and increased backward slip rather than excessive sinkage. By extending the model to generalized terrain conditions, we construct failure phase diagrams that identify sinkage- and slippage-induced failure regimes, enabling quantitative risk estimation for locomotion on granular slopes. This physics-informed framework provides predictive insight into terrain-dependent failure mechanisms and offers guidance for safer and more robust robot operation on deformable inclines.

\end{abstract}

\section{Introduction}\label{sec:Intro} 

\begin{figure}[htbp!]
\centering
\includegraphics[scale=0.52]{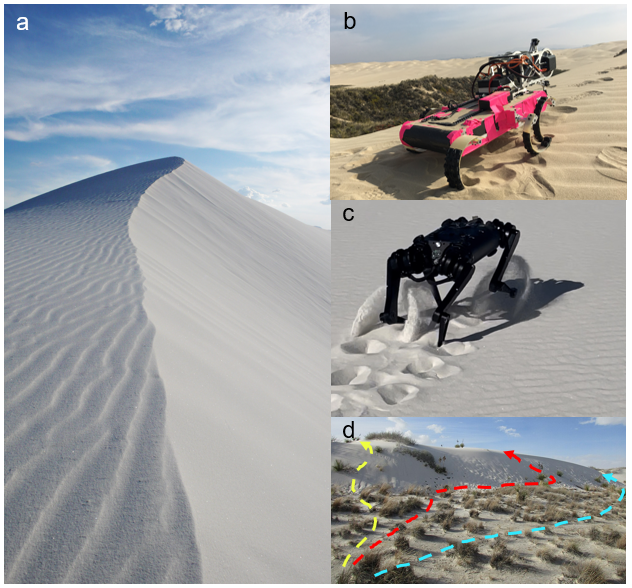}
\caption{Importance and challenges of legged locomotion on sand slopes.
{\textbf{(a)}} Sand slopes widely present in natural environments (White Sands National Park, NM).
{\textbf{(b)}} Hexapedal robot ascending a dune slope.
{\textbf{(c)}} Quadruped robot traversing a relatively mild sand slope at White Sands.
{\textbf{(d)}} Example ascent paths on a sand slope, illustrating the tradeoff between shorter routes (yellow) and longer but potentially less risky path (red, cyan).
}

\label{Fig.Introduction}
\end{figure}

Sand slopes are widespread in natural environments, including deserts, coastal regions, riverbanks, and planetary surfaces (Fig. \ref{Fig.Introduction}a). These terrains are of increasing interest for mobile robots, for applications such as  search-and-rescue, environmental monitoring, rural delivery, and extraterrestrial exploration. However, the combination of deformable substrates and steep inclines in these environments pose significant challenges to terrestrial locomotion, where robots often encounter excessive sinkage, slippage, or loss of stability~\cite{hu2025learning,li2013terradynamics,qian2015dynamics} (Fig. \ref{Fig.Introduction}b, c). Misapplied locomotion strategies can lead to catastrophic failures and complete immobilization. A mechanistic understanding of how robots interact with sand slopes is therefore essential for assessing traversal risks and extending operational envelopes~\cite{sorice2021insight} (Fig. \ref{Fig.Introduction}d). 

A fundamental challenge of locomotion on sand is that granular substrates can exhibit both solid-like and fluid-like behaviors~\cite{jaeger1992physics,nedderman1992statics}, leading to complex interactions between limb intrusion, force generation, and terrain yielding~\cite{bekker1956theory,wong1989terramechanics,chen2025dynamic,chen2023energy}. 
Prior studies of locomotion on level granular media have characterized resistive forces during intrusion and propulsion, leading to models that successfully predict performance on level sand~\cite{li2009sensitive,qian2015principles}. These works established how penetration depth and normal resistive forces regulated step length and speed on level sand. 



On inclined sand, however, gravity alters the internal stress state of the granular medium, reducing available shear resistance and promoting downslope yielding~\cite{pouliquen2002friction,jop2006constitutive}. Previous robotic and biological studies have shown that increasing slope angle degrades speed, increases energetic cost, and alters gait stability~\cite{marvi2014sidewinding,gosyne2018bipedial,hu2025learning}. Bio-inspired strategies, such as lateral undulation in snakes or specialized limb kinematics in insects, have demonstrated improved performance by mitigating slip and redistributing forces~\cite{humeau2019locomotion,marvi2014sidewinding}. Together, these studies highlight the importance of force generation and terrain yielding in determining locomotor success on granular inclines. Despite this progress, most existing research on robot locomotion on slopes~\cite{kolvenbach2022traversing,roberts2016rhex,hu2025learning} do not directly resolve the underlying force interactions between the robot and the granular medium on slopes. 
Without linking performance degradation to measurable changes in normal and shear resistance, it remains difficult to determine whether failures arise primarily from excessive sinkage, insufficient shear anchoring, or their interaction. Consequently, predictive understanding of slope-induced failure mechanisms remains limited.

To address this gap, in this paper we experimentally measured the locomotion performance of a hexapedal robot on a tiltable sand fluidization bed (Sec. \ref{sec:Design}). By systematically quantifying slope-dependent normal and shear resistances, we identified how granular anisotropy alters limb anchoring dynamics. We then developed a simplified force-balance model linking shear strength to anchoring time, step length, and speed (Sec.~\ref{sec:results}). Finally, we generalized this framework to construct terrain–robot phase diagrams that distinguish sinkage- and slip-dominated failure regimes, enabling model-informed risk prediction for locomotion on granular slopes (Sec.~\ref{sec:General Shape}).

\section{Materials and Methods}\label{sec:Design}
\subsection{Robot} 
\begin{figure}[htbp]
\centering
\includegraphics[scale=0.7]{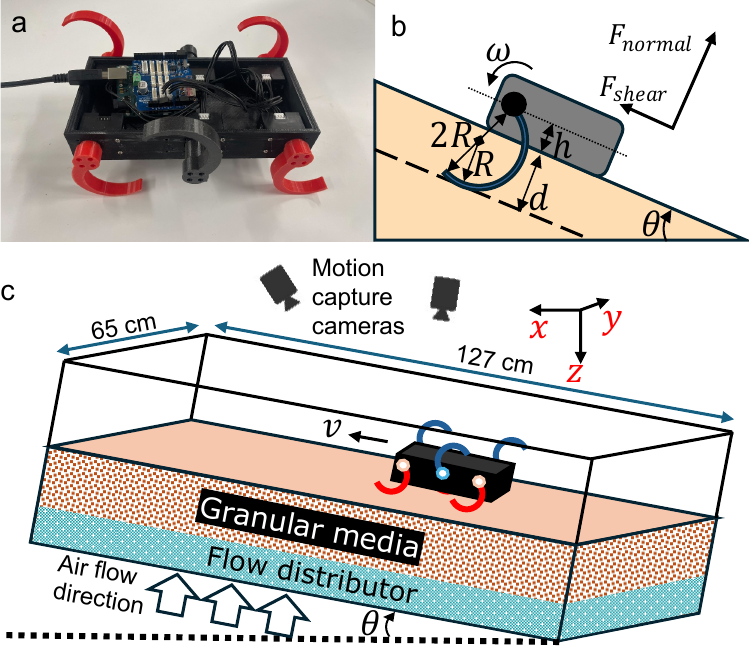}
\caption{Robot and experimental setup for studying locomotion on granular slopes. {\textbf{(a)}} Hexapedal robot used in this study. {\textbf{(b)}} Diagram illustrating the key parameters of robot leg. Here $R$ denotes robot leg radius, $d$ denotes the leg penetration depth, $\omega$ denotes the leg angular velocity, and $h$ denotes the hip height, measured as the distance from the leg axle to the bottom of the robot body.  {\textbf{(c)}}Schematic of the tiltable granular trackway for testing robot locomotion on sand. In b and c, $\theta$ denotes the slope angle of the granular incline.}
\label{Fig.structure}
\end{figure}

To investigate locomotion performance on sand slopes, we developed a simple hexapedal robot (22 cm length $\times$ 10 cm width, 350 grams,  Fig. \ref{Fig.structure}a) to perform controlled locomotion experiments. 
The sand-climbing hexapod robot were equipped with six 3-D printed C-shaped legs (Fig. \ref{Fig.structure}a,b). The angular positions of the robot legs  was controlled using a microcontroller (Arduino Uno), and actuated by servo motors (Dynamixel XL330). The radius of the C-shaped leg was 4 cm and the width of the leg was 1 cm. 
During all trials, the robot was programmed to use an alternating tripod gait~\cite{saranli2001rhex} in which the legs were divided into 2 groups (front and rear leg from one side, plus the middle leg from the other side), and the two groups alternated with a 180-degree phase difference with each other. This alternating tripod gait is a bio-inspired gait often used by insects~\cite{cham2004stride}, and was found to exhibit good locomotion performance on leveled sand~\cite{humeau2019locomotion,li2009sensitive}. 
For all trials, the leg angular speed was kept constant at $\omega=2\pi$ rad/s. 

\subsection{Robot Locomotion Experiments} \label{sec:setup}
Robot locomotion experiments were performed in a 127 cm $\times$ 65 cm trackway (Fig. \ref{Fig.structure}c). The trackway was filled with 
300 $\mu$m diameter glass beads (Grainger) as a model granular substrate, to a depth of 10 cm to minimize boundary effects~\cite{li2013terradynamics}. The glass beads have been proven to be a promising model substrate for studying the interaction between robots and granular materials~\cite{mueth1998force}, while the simpler geometry facilitates the control of the substrate to achieve uniform packing state~\cite{nayak2025burrowing}. To ensure a consistent initial condition across  locomotion tests, the granular bed was fluidized prior to each trial by sending air upward through the base of the trackway. After the air flow slowly ramped down, the sand would settle into a uniform, loosely-packed compaction, eliminating spatial variations in surface height and penetration resistance~\cite{jin2019preparation}. 
The granular slope angle, $\theta$, was controlled using two linear actuators, and measured using a digital angle measurer.
The maximum tilting angle used in the experiment was 24 degrees, near the angle of repose~\cite{elekes2021expression,dai2016observed} of the granular media used.
To characterize the robot speed while traveling on sand, three motion capture cameras (Optitrack, Prime 13W) were installed around the trackway to record the position of the center of mass (CoM) of the robot in the world frame ($x$; $y$; $z$) 
at 120 frames per second (FPS). In addition, two video camera (Optitrack Prime Color, and GoPro HERO 12) were used to record experiment footage from top and side view at 120 FPS and 30 FPS, respectively.

\subsection{Penetration and Shear Sand Resistive Force Measurements}
To investigate how sand inclination alters internal force transmission, we quantified granular resistive forces in both the normal direction (perpendicular to the sand surface) and the shear direction (parallel to
the surface) as a function of slope angle. 

A belt-driven linear actuator (30 cm travel) powered by a NEMA 23 stepper motor was mounted above the tiltable fluidized bed. A miniature load cell (HX711 amplifier, Arduino Uno for data acquisition) was installed in-line between the actuator end-effector and the intruding plate to directly measure penetration and shear forces. The setup was modular, allowing rapid reconfiguration between normal penetration and tangential shear experiments (Fig. \ref{Fig.sand force}a, e).

Measurements were conducted at six inclination angles\footnote{Inclinations above $24^\circ$ were not tested because $25^\circ$ approaches the angle of repose of the prepared sand, beyond which the surface becomes unstable and prone to spontaneous yielding.}: \(0^\circ, 5^\circ, 10^\circ, 15^\circ, 20^\circ,\) and \(24^\circ\). 
Before each trial, the granular bed was fully fluidized by passing air upward through the base of the trackway to reset the packing state. After fluidization, airflow was stopped and the bed was slowly tilted to the commanded angle prior to intrusion.

In penetration trials, a circular disk (radius = 1 cm) was driven into the sand at a constant speed of 5 cm/s along the direction normal to the inclined surface (Fig. \ref{Fig.sand force}b). The load cell recorded the
resistive force as a function of penetration depth, enabling characterization of slope-dependent normal resistance.

In shear trials, a rectangular plate (height = 2 cm, width = 3 cm) was inserted fully into the sand and
translated down-slope at 5 cm/s parallel to the surface (Fig. \ref{Fig.sand force}f). Prior to each experiment, calibration trials were conducted in air at each slope angle to determine force offsets. The maximum penetration depth of the shear plate was maintained at 2 cm for all trials. During translation,
the load cell recorded the tangential resistive force exerted by the sand, allowing quantification of slope-dependent shear resistance.

\section{Results and discussion}\label{sec:results} 

\subsection{Effect of inclination angle on sand climbing performance}\label{sec:performance}  

\begin{figure*}
\centering
\includegraphics[width=1\linewidth]{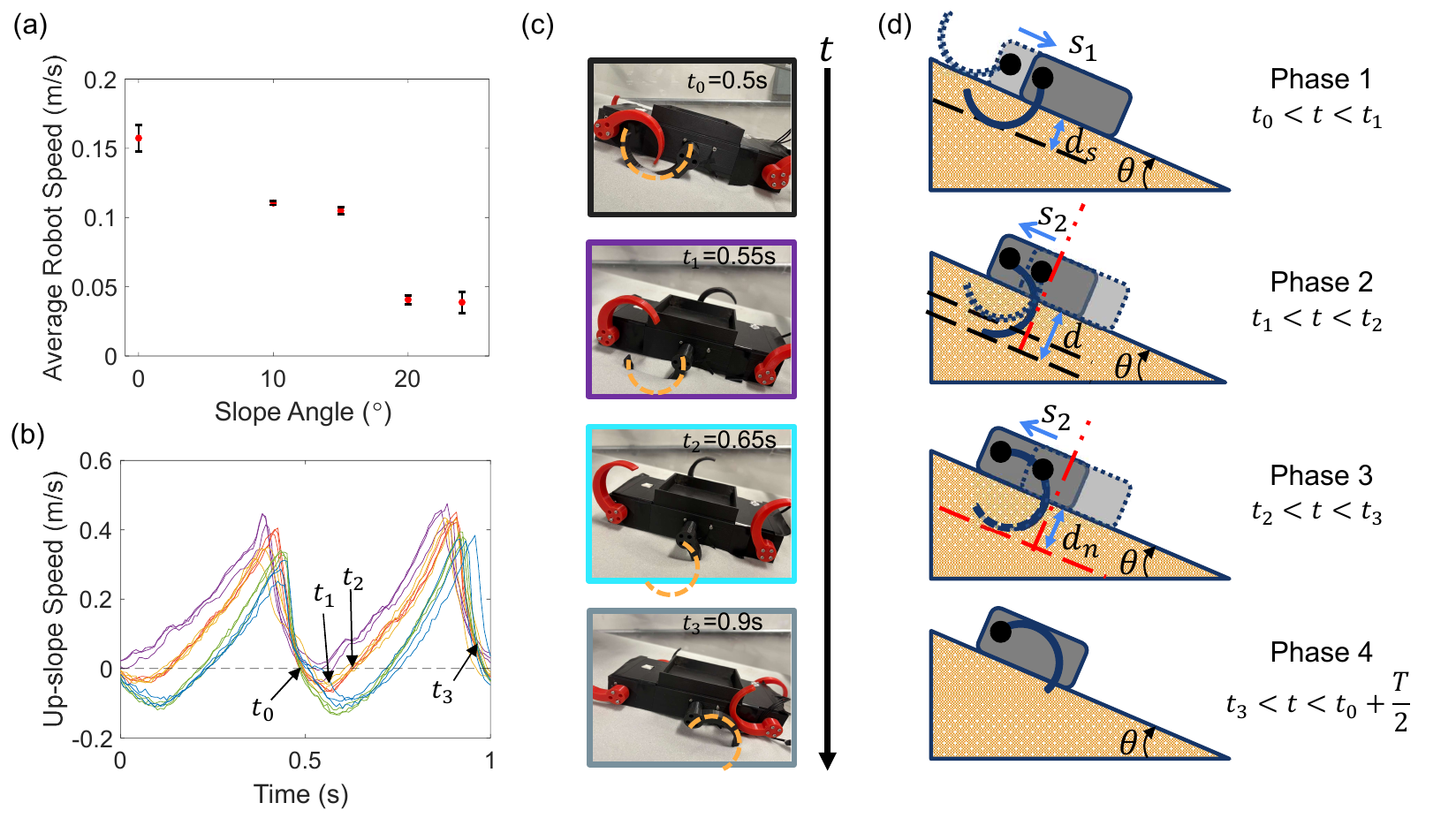}
\caption{Robot locomotion on sand slopes of varying inclinations. 
{\textbf{(a)}} Average robot up-slope speed as a function of slope angle. 
{\textbf{(b)}} Experimentally-measured robot speed vs. time over one stride period $T$. Purple, red, orange, green, and blue curves represent 0, 10, 15, 20, 24 degree inclinations, respectively. 
{\textbf{(c)}} Experimental image sequences showing the side view of the robot traveling on a 15-degree slope. Corresponding critical moments, $t_0$,$t_1$, $t_2$, $t_3$ are also labeled on the orange (15-deg) curve in (b).  
{\textbf{(d)}} Schematic of the leg-ground interaction during each phases. 
In c and d, images from up to bottom correspond to the critical timings, $t_0$, $t_1$, $t_2$, and $t_3$, labeled in b.
} 
\label{Fig.results}
\end{figure*}
Tracked robot speed on sand slopes (Fig. \ref{Fig.results}a) showed that among the 5 tested inclinations, $\theta=(0^\circ,10^\circ,15^\circ,20^\circ,24^\circ)$, the robot on level sand at 0 degrees exhibited the fastest forward speed ($\bar{v} $ = 15.7 $\pm$ 0.97 cm/s). At moderate slopes of 10$\degree$ and 15$\degree$, the robot speed decreased by approximately 30\% ($\bar{v}$ = 11.0 $\pm$ 0.14 cm/s at 10 degrees, and $\bar{v}$ = 10.5 $\pm$ 0.23 cm/s at 15 degrees). On steep granular slopes with inclination angle exceeding 20$\degree$, robot speed decreased to nearly one third of level-sand value ($\bar{v}$ = 4.98 $\pm$ 0.32 cm/s at 20 degrees, and  $\bar{v}$ = 5.6 $\pm$ 1.6 cm/s at 24 degrees). Here $\bar{v}$ is the step-averaged robot velocity along the direction parallel to the sand surface. 

Inspection of the instantaneous speed $v$ within a stride (Fig.~\ref{Fig.results}b) revealed a consistent temporal structure in every trial. Touchdown of the stance legs occurred at $t_0=0$ s (and again at $t_0=0.5$ s for the alternating leg group), when the legs began penetrating the sand surface. On level sand, robot speed increased immediately after touchdown (Fig. \ref{Fig.results}b, purple curve). In contrast, as inclination angle increased (Fig. \ref{Fig.results}b, red, orange, green, blue curves), the robot consistently exhibited a backward slip following touchdown. During this interval, the instantaneous speed became negative (Fig. \ref{Fig.results}b, phase 1, $t_0$ to $t_1$), indicating down-slope motion of the body relative to the ground.

We noticed that the anchoring timing $t_1$, the instant when backward slippage stopped accelerating (i.e. when $\dv {v(t)}{t}$ increased from negative to positive values), was highly repeatable across trials at the same inclination, and increased monotonously with the increasing slope angle. 
After anchoring, the robot enters a propulsion phase, during which the upslope speed increases at a rate that is largely insensitive to inclination (Fig. \ref{Fig.results}b, phase 2, 3, $t_1$ to $t_3$). This observation suggested that degradation in climbing performance was not primarily caused by reduced propulsive capability, but rather by delayed effective anchoring timing.
Finally, as the stance legs extracted from the sand at lift off timing $t_3$ , the robot speed decreased toward zero, as the legs transitioned into the aerial phase (Fig. \ref{Fig.results}c, d, phase 4) until the alternating leg group makes contact and the next stride began.

According to previous literature, the anchoring timing on level sand was governed by the penetration depth~\cite{li2009sensitive,qian2015principles}.
In granular media, the force resisting leg penetration increases linearly with depth:  $F_n = k_n dA$, where $k_n$ is the normal direction penetration resistance per unit contact area, $d$ is the leg penetration depth, and A is the contact area. During early contact, the applied normal force exceeds the resistive force the substrate can provide, and the sand yields locally. In this regime, the robot leg ``swims'' through the fluidized sand and fails to anchor. 
As penetration increases, the resistive force grows until reaching a normal equilibrium depth $d_n$, where the sand can support the applied load and locally solidifies. At this point, the leg can generate effective thrust and initiate forward propulsion (Fig. \ref{Fig.results}c, d, phase 3). In the rotary walking model~\cite{li2009sensitive}, the step length during this propulsion phase is given as $s=2\sqrt{R^2-(d_n+h-R)^2}$, where $R$ is the C-leg radius, $h$ is the hip height, and $d_n$ is the sand solidification depth (Fig. \ref{Fig.structure}b). 

Motivated by this framework, we investigated whether the observed performance degradation on slopes would arise from increased sinkage. We therefore formulated \textbf{Hypothesis 1:} As slope angle increased, the normal-direction sand penetration resistance decreased, resulting in greater sand solidification depth, $d_n$, and consequently reduced step length during the propulsion phase, $s_2$. To test this hypothesis, in the next section we systematically quantified the normal penetration resistance of sand as a function of slope angle, and evaluated the predicted displacement against experimental observations. 

\subsection{Sinkage-induce locomotion performance loss on granular slope}\label{sec:sinkage}

\begin{figure*}[bthp!]
\centering
\includegraphics[width=1\linewidth]{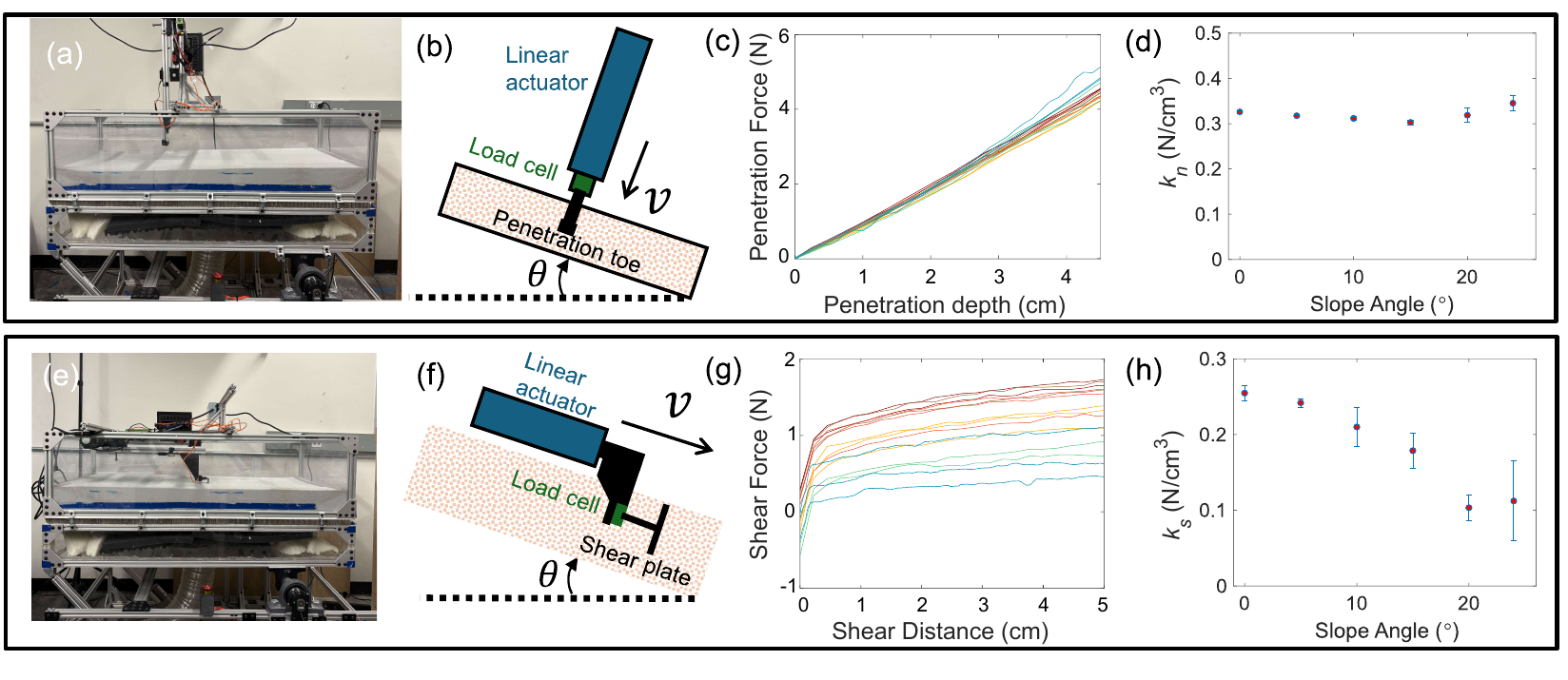}  
\caption{Sand resistive force measurements on inclined granular surfaces. 
{\textbf{(a, e)}} Experimental setups for penetration and shear force measurements. 
{\textbf{(b, f)}} Schematics illustrating the penetration and shear plate motion. 
{\textbf{(c, g)}} Measured penetration and shear resistive forces, as functions of penetration and shear displacements. Red to blue colors represent inclination angle from 0$\degree$ to 24$\degree$. 
{\textbf{(d, h)}} Measured sand penetration and shear resistance per unit area as functions of slope angles.  
}
\label{Fig.sand force}
\end{figure*}

To test Hypothesis 1 we performed controlled penetration trials on slopes of varying inclinations (Fig.~\ref{Fig.sand force}a, b). The measured normal force vs penetration depth was approximately linear for all inclinations tested (Fig.~\ref{Fig.sand force}c). A linear regression of the $F_n$ vs. $d$ curves yielded the effective normal penetration resistance per unit contact area, $k_n$. Surprisingly, $k_n$ showed negligible change across the range of slope angles we examined (Fig. \ref{Fig.sand force}d). 

As a result, using the measured $k_n$ to compute the equilibrium normal depth $d_n$ and the resulting step length, $s_2=2\sqrt{R^2-(d_n+h-R)^2}$, produced only minimal variation with slope. Thus, although the previously established rotary-walking mechanism correctly linked $d_n$ to step length on level sand, our measurements indicated that changes in normal penetration resistance are too small across the tested inclinations to explain the large, observed reduction in climbing speed. In short, Hypothesis 1 was not supported by the data: sinkage-induced effects appeared unlikely to be the primary driver of the delayed anchoring and speed loss on slopes.

Consistent with this conclusion, leg–terrain interaction videos showed comparable maximum penetration depth across inclinations but substantially increased backward slip on steeper slopes. Based on this observation, we suspected that on sloped sand, the slippage, rather than excess sinkage, was the dominant effect on robot locomotion performance. Granular-physics studies provided a mechanism for this behavior. In the shear direction, local yielding occurred when the applied shear stress exceeded the sand’s local shear yield strength. Yielding was accompanied by force-chain breakage and grain rearrangement, resulting in loss of rigid support beneath the intruder~\cite{mueth1998force,peters2005characterization,switala2024micro,liao2025bio}. For a robot climbing a slope, this local weakening allowed the downslope gravitational component to exceed the resisting strength, producing slippage (and in extreme cases, avalanching)~\cite{barker2000two}.

Because shear resistance force increases with penetration depth~\cite{Sidewinding}, we hypothesized that at initial shallow contact the applied shear load exceeded the local shear yield strength, causing inevitable downslope slip (Fig.~\ref{Fig.results}c, d, phase 1). Only after the leg penetrated sufficiently to reach a shear-equilibrium depth, $d_s$, where applied shear equaled resisting shear, would the substrate locally solidify behind the leg and allowed effective anchoring and propulsion, as if the leg was rotating against a vertical support normal to the sand surface (Fig. \ref{Fig.results}c, d, phase 2). 

We therefore formulated \textbf{Hypothesis 2}: As slope angle increased, the shear-direction sand resistance decreased, which increased the shear equilibrium depth, $d_s$, thereby increasing backward slip $s_1$ during the intrusion phase and reducing net step displacement, $s=s2-s1$. To test this hypothesis, in next sections we systematically quantified the shear resistance of sand as a function of slope angle, and evaluated the model-predicted displacement, $s_2 - s_1$, against experimental observation. 


\subsection{Slippage-induce locomotion performance loss on granular slope}\label{sec:slippage}

To test Hypothesis 2, we performed controlled shear experiments on sand slopes of varying inclination to quantify the shear resistance and estimate the shear-equilibrium depth, $d_s$. Analogous to the normal penetration depth, $d_n$, we assume that the shear-equilibrium depth depends on the sand’s effective shear strength, $k_s$, which may vary with slope angle, $\theta$. Thus, $d_s$ can be expressed implicitly as a function of $k_s (\theta)$ through shear-force equilibrium.

The measured shear force–displacement curves are shown in Fig.~\ref{Fig.sand force}g. In all trials, the shear force rises rapidly at small shear distances and then transitions into a steady state regime. Although the qualitative behavior is consistent across inclinations, the overall magnitude of the shear force decreases systematically as slope angle increases.

This two-phase response agrees with prior granular-physics studies, which suggested that the initial rapid rise in shear force corresponds to dynamic grain rearrangement and compression, while the subsequent slower-growth regime corresponds to the formation of a stable force-chain network or frictional shear band within the granular bed~\cite{li2013terradynamics,zhang2014effectiveness,albert1999slow,da2005rheophysics}. We therefore define the effective shear resistance using the average value of the quasi-static plateau region. For a plate-like intruder spanning from the surface to depth, $d$, the steady-state shear resistance scales with the square of depth, $F_s\propto d^2$~\cite{roth2021constant,zhang2014effectiveness}. Accordingly, the shear resistive force can be modeled as $F_s=k_sWd_s^2$, where $W$ is the width of the leg, and $k_s$ characterizes the shear strength of the granular substrate. 

Fig. \ref{Fig.sand force}h shows the average shear resistance force as a function of slope angle. In contrast to the normal-direction resistance, $k_n$, which remains nearly constant, the shear resistance decreases markedly with increasing inclination. For example, the average shear force measured at $20\degree$ is only approximately 40\% of that measured on level sand. 
This reduction in shear strength has two important implications. First, because the applied downslope shear load increases with $\sin(\theta)$ while the resisting strength decreases, the shear-equilibrium depth, $d_s$, must increase with slope angle. Second, since penetration depth grows monotonically during leg rotation in phase 1, a larger $d_s$ requires a longer time to reach shear equilibrium. Since the robot leg cannot secure the anchoring in sand and propel the body forward before it reaches the shear equilibrium depth, we believe that the observed anchoring timing $t_1$ closely resembles the timing that the leg reaches the shear equilibrium such that $d(t_1) \approx d_s$. This directly explains the delayed anchoring timing, $t_1$, observed in Sec.~\ref{sec:performance}, as a result of decreased $k_s$. 

In the following section, we formalize this relationship by constructing an interaction model that uses the experimentally measured shear strength $k_s (\theta)$ to predict the anchoring timing, $t_1$, and, consequently, the robot’s step length and speed on granular slopes.
 
\subsection{Hypothesized mechanism governing robot critical timings on granular slope}\label{sec:hypothesis}
In this section, we formalize the leg–sand interaction into a predictive model that links anchoring timing $t_1$ to slope angle $\theta$ through the measured sand shear strength $k_s$. 

During phase 1, the robot applies a downslope force to the sand:  $m(g+a)\cdot\sin(\theta)$. Here $mg\cdot\sin(\theta)$ represents the horizontal component of the gravitational force due to robot weight, and the acceleration $a$ can be expressed as $R\omega/\Delta t$, where $\Delta t=0.2s$ represents the characteristic elastic response time~\cite{li2009sensitive}. In the alternating tripod gait, the load is distributed across $n=3$ stance legs, and thus the average applied shear load per leg is
\begin{align*} 
& F_{a}=\frac{m(g\cdot \sin(\theta)+\frac{R\omega}{\Delta t})}{n} \nonumber
\end{align*}
Meanwhile, the resistive shear force provided by the sand increases quadratically with penetration depth (Sec.~\ref{sec:slippage}): 
\begin{align*} 
& F_{s}=k_s(\theta) W d_s^2 \nonumber
\end{align*}
Once applied and resisting shear forces balance, \ie $F_{a}=F_{s}$, sand solidifies in the shear direction, and the anchoring phase begins. The shear-equilibrium depth, $d_s$, can therefore be determined:
\begin{align} 
& d_s=\sqrt{\frac{m(g\cdot \sin(\theta)+\frac{R\omega}{\Delta t})}{nk_s(\theta) W}} 
\end{align}
Because $k_s(\theta)$ decreases with slope angle while the applied shear load increases with $\sin\theta$, this expression predicts a monotonic increase of  $d_s$ with $\theta$, consistent with the experimentally-observed delay in anchoring. 

We subsequently compute the anchoring time, $t_1$, from the shear-direction sand solidification depth, $d_s$. Given the C-leg geometry (Fig.~\ref{Fig.structure}b), the leg penetration depth during stance can be expressed as a function of time: 
\begin{align}
    d(t) = R\sin(\phi(t) +\phi_0)+R-h.
    \label{eqn:penetration-depth}
\end{align}
Here $R$ is the leg radius, $\phi(t) = \omega t$ is the leg angular position at time $t$, $\phi_0 =\arcsin \frac{h-R}{R}$ is the touchdown angle at $t_0$, and $h$ is the hip height (Fig. \ref{Fig.structure}b). 

Using Eqn. \ref{eqn:penetration-depth}, the anchoring timing $t_1$, at which the leg reaches the shear equilibrium depth, can be determined from $d_s$: 
\begin{align} 
& t_1=\frac{\arcsin(\frac{d_s+h}{R}-1)-\phi_0}{\omega} \label{eq:myalign}
\end{align}
This expression makes explicit that the anchoring timing is governed by the shear-equilibrium depth $d_s$. As slope angle increases, the increased $d_s$ will result in the delayed $t_1$.

\begin{figure}[htbp]
\centering
\includegraphics[width=0.7\linewidth]{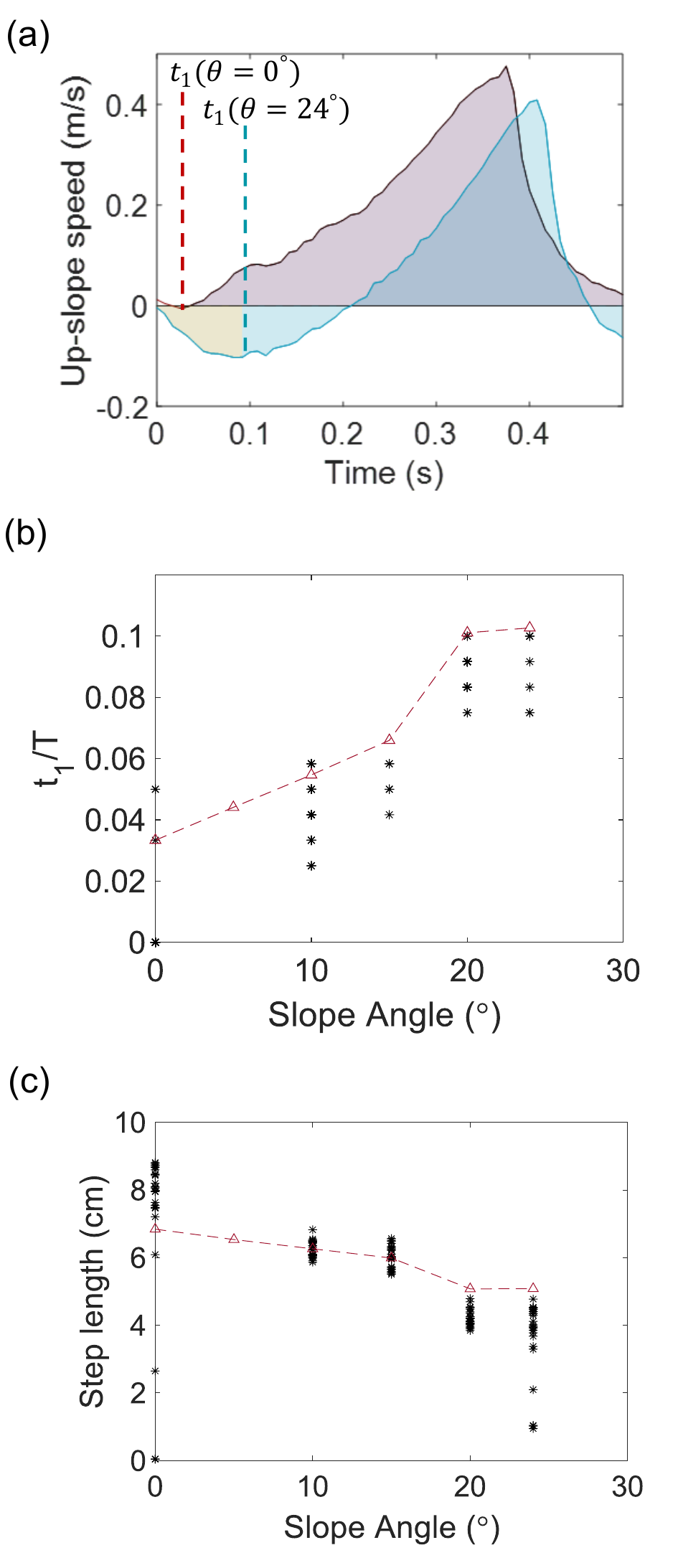}
\caption{Predictive model to link anchoring timing to robot step length.  \textbf{(a)} Comparison of robot speed vs. time on 0$\degree$ and 24$\degree$ sand slope in one representative step. Red and blue vertical dashed lines denote the anchoring timing $t_1$ for 0$^\circ$ and 24$^\circ$, respectively. The yellow shaded region indicates the backward slippage $s_1$ on the 24$^\circ$ slope. The purple and blue shaded regions (net positive area under the velocity curve) denote the propulsion displacement $s_2$ for 0$^\circ$ and 24$^\circ$, respectively.
\textbf{(b)} Experimentally-measured (black asterisks) and model-predicted (red dashed line) $t_1$ as a function of slope angle. 
\textbf{(c)} Experimentally-measured (black asterisks) and model-predicted (red dashed line) robot net step length $s$ as a function of sand slope angle. 
}
\label{Fig.failure mechanism}
\end{figure}

Model-predicted $t_1$ values (Fig. \ref{Fig.failure mechanism}b, red dashed line) successfully captured both the monotonic increasing trend and the magnitude of the experimental measured anchoring time (Fig. \ref{Fig.failure mechanism}b, black asterisks). Since the rate of up-slope velocity increase is nearly identical in Phases 2 and 3, the effective propulsion dynamics appear to be only governed by $t_1$, supporting Hypothesis 2. In the next section, we translate the predicted anchoring timing into a closed-form estimate of robot step length and speed.

\subsection{Model-predicted robot step-length on sand slope}\label{sec:mechanism}
In Sec. III-D, we showed that slope angle modifies shear strength $k_s(\theta)$, which determines the shear-equilibrium depth $d_s$, and therefore the anchoring timing $t_1$. We now quantify how this delayed anchoring reduces the robot’s step length.

To calculate the total slippage, we assumed that upon touchdown, the robot slipped downhill under the  gravitational component $mg\cdot \sin(\theta)$, until the leg was able to solidify the sand in the shear direction at the anchoring timing $t_1$. The total slippage before sand solidification (Fig. \ref{Fig.failure mechanism}a, yellow shaded region), can therefore be approximated as:
\begin{align} 
s_1 =\iint_{t_0}^{t_1} g\cdot \sin\theta \,\mathrm{d}t\,\mathrm{d}t=\frac{1}{2}g\sin\theta\cdot{t_1}^2.
\end{align}
Thus, slippage increases quadratically with anchoring delay.

After $t_1$, the leg began to push against solidified sand and generate effective propulsion. Accordingly to our experimentally-measurements (Fig.~\ref{Fig.results}b), once anchoring occurs, the upslope acceleration during propulsion was primarily geometric based (\ie leg pushing against a solid side wall, Fig. \ref{Fig.results}d, phase 2) and independent of slope angle. As a result, the displacement during propulsion phase, $s_2$ (Fig. \ref{Fig.failure mechanism}a, net positive area of the blue shaded region), can be approximated based on the anchoring timing, $t_1$:
\begin{align} 
&s_2= s^*(\frac{T/2-t_1}{T/2})^2-g\sin\theta\cdot t_1(T/2-t_1),
\end{align}
where $s^*$ denotes the step length on level sand (Fig. \ref{Fig.failure mechanism}a, red shaded region), $T/2$ represents propulsion duration on level sand, and $T/2-t_1$ represents the shortened propulsion duration on sloped sand. The first term captures reduced propulsion time; the second accounts for inherited initial speed at $t_1$ because of slippage. 

The total net displacement per step, therefore, can be expressed as 
\begin{align} 
& s = s_2 - s_1
= -\frac{1}{2}g\sin\theta\cdot t_1(T-t_1)+s^*(\frac{T/2-t_1}{T/2})^2. 
\end{align} 
   
Theoretical predictions of robot step length $s$ (Fig.~\ref{Fig.failure mechanism}c, red dashed line) agreed well with the experimentally measured values (black asterisks), confirming that slope-dependent reductions in shear strength and the resulting anchoring delay governed the observed performance loss. 

\section{Model generalization and broader applicability}\label{sec:General Shape}

\begin{figure}[bthp!]
\centering
\includegraphics[width=1\linewidth]{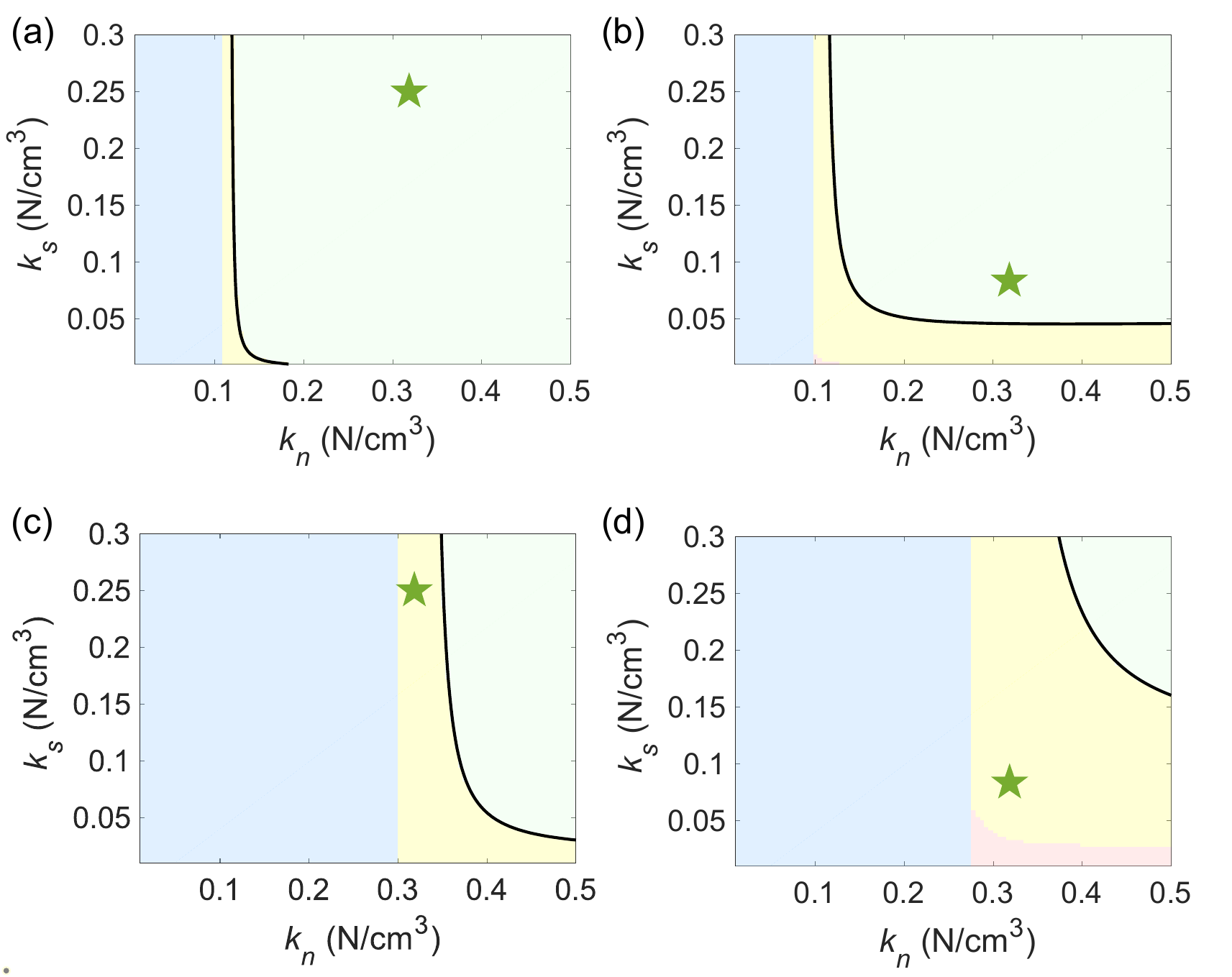}  
\caption{
Model-predicted robot performance on (a) level sand when the total mass $m = 0.35$kg, (b) a $24^\circ$ sand slope when $m = 0.35$kg, (c) level sand when equipped with a heavier load that $m = 1$kg, and (d) a $24^\circ$ sand slope when $m = 1$kg. Blue regions denote sinkage-induced failure, red regions indicate total slippage failure, and yellow regions represent the metastable cases that experience potential locomotion risks. The light green regions correspond to successful locomotion. The black lines are the step length contours where $s=R$ which show the boundary of the success region. The dark green stars mark the experimentally measured sand resistances reported in Fig.~\ref{Fig.sand force}.}
\label{Fig.Generalized model}
\end{figure}

In this section, we extend the experimentally validated shear–anchoring model into a predictive framework for locomotion risk across a broad range of terrain and robot parameters. Rather than restricting analysis to the specific sand conditions tested experimentally, the model enables the generation of performance phase diagrams (Fig.~\ref{Fig.Generalized model}) that predict whether locomotion succeeds, fails by sinkage, or fails by slippage, given the terrain’s normal and shear resistances and the robot’s weight.

Each map partitions terrain space into distinct locomotion regimes. \textbf{Sinkage-induced failure} (Fig.~\ref{Fig.Generalized model}, blue regions) occurs when normal resistance $k_n$ is insufficient to support the applied load. Excessive penetration prevents formation of a stable pivot, causing the leg to “swim” through fluidized sand and produce negligible propulsion. \textbf{Slippage-induced failure} (Fig.~\ref{Fig.Generalized model}, red regions) arises when shear resistance $k_s$ is too low relative to the applied downslope shear load. In this regime, the predicted net step length $\tilde{s}=\tilde{s_2}-\tilde{s_1}$ becomes negative, indicating downslope slippage. \textbf{Metastable regimes} (Fig.~\ref{Fig.Generalized model}, yellow regions) correspond to small predicted step length ($0<s<R$), where the robot advances less than one leg radius per step and therefore places its next foothold within previously disturbed sand, and each successive step would experience diminished support\cite{li2009sensitive,liu2023adaptation}, leading to progressively smaller step length and eventual transition to either sinkage or slippage dominated failure during sustained operation. The green region denotes robust locomotion ($s>R$), where sustained propulsion can be achieved. 

The phase diagrams revealed several important insights. First, slope inclination primarily expands the slippage-dominated regime through modulation of $k_s$, dramatically shrinking the feasible locomotion region (compare Fig.~\ref{Fig.Generalized model}a \vs b). In addition, increasing robot mass shifts both failure boundaries (Fig.~\ref{Fig.Generalized model}c, d). Under the terrain strength used in our experimental study (green star), the model predicts that a 1-kg robot would encounter metastable regime on a $24^\circ$ slope, and eventually fail due to slippage-dominated failure (Fig.~\ref{Fig.Generalized model}d).

These maps demonstrate that locomotion performance on granular slopes is not governed by a single failure mode, but by the interaction between normal support and shear anchoring. By explicitly separating these mechanisms, the model provides a quantitative tool for predicting how terrain mechanics and robot design jointly determine success or failure. Beyond explanation, this risk-estimation framework supports terrain-aware path planning and robot design optimization. By estimating slippage risk and anchoring delay along different routes, the robot can select safer paths rather than simply the shortest ones. In addition, it can inform the selection of robot weight, leg size, and gait parameters, to ensure safe traversal of sand slopes with a wide range of strength.

\section{CONCLUSION}
In this work, we investigated the mechanisms governing legged robot locomotion on granular slopes by combining controlled experiments with a simplified force-balance model. Through systematic measurements of slope-dependent normal and shear resistances, we showed that performance degradation on inclined sand is driven primarily by reduced shear strength rather than increased normal sinkage. The decrease in shear resistance delays anchoring, increases backward slip, and shortens the effective propulsion phase within each step, leading to reduced step length and speed. Building on this mechanism, we developed an analytical robot-sand interaction model that quantitatively links the sand shear strength to robot leg anchoring time and step length, across a wide range of slope angles. We further showed that the model enables construction of failure phase diagrams that distinguish sinkage-, slippage-, and metastable regimes, allowing locomotion risks to be assessed beyond the specific terrain conditions tested experimentally. Overall, our results identify the physical origin of slope-induced failure on granular media and provide a predictive framework for estimating locomotion risks across terrain conditions. This physics-grounded approach offers quantitative guidance for terrain-aware planning and safe operation on deformable slopes.

\bibliographystyle{./bibliography/IEEEtran.bst}
\bibliography{./bibliography/terradynamics.bib, ./bibliography/mudforce.bib, ./bibliography/mudsensing.bib, bibliography/qian, bibliography/rolling, bibliography/slope.bib}


\clearpage

\end{document}